\documentclass[10pt,twocolumn,letterpaper]{article}

\usepackage[pagenumbers]{cvpr} % To force page numbers, e.g. for an arXiv version
\usepackage{tikz}

\usepackage{tcolorbox}
\usepackage{colortbl}
\usepackage{mdframed}
\usepackage{bbm}
\usepackage{bm}
\mdfsetup{%
linecolor=white,
backgroundcolor=gray!10,
}
\usepackage{amssymb}% http://ctan.org/pkg/amssymb
\usepackage{pifont}% http://ctan.org/pkg/pifont
\newcommand{\cmark}{\ding{51}}%
\newcommand{\xmark}{\ding{55}}%
\newcommand{\norm}[1]{\left\lVert#1\right\rVert}

\definecolor{cvprblue}{rgb}{0.21,0.49,0.74}
\definecolor{Gray}{gray}{0.9}
\definecolor{palegray}{HTML}{F5F5F5}
\definecolor{barred}{HTML}{F24B4B}
\definecolor{barblue}{HTML}{1DB6F2}
\definecolor{baryellow}{HTML}{F2A922}
\definecolor{barpurpule}{HTML}{A441BF}
\definecolor{bargreen}{HTML}{97BF5A}
\definecolor{palepurpule}{HTML}{FFF8FF}
\definecolor{palered}{HTML}{FFCCCB}
\definecolor{paleblue}{HTML}{CAF7F7}
\definecolor{palegreen}{HTML}{F6FAF3}
\usepackage[pagebackref,breaklinks,colorlinks,citecolor=cvprblue]{hyperref}

\newcommand{\ours}{CompFuser}

\def\paperID{8536} % *** Enter the Paper ID here
\def\confName{CVPR}
\def\confYear{2024}

\title{Unlocking Spatial Comprehension in Text-to-Image Diffusion Models}
\author{%
\hspace{4mm}Mohammad Mahdi Derakhshani\textsuperscript{*}%
\hspace{8mm} Menglin Xia\textsuperscript{\dag}%
\hspace{7mm} Harkirat Behl\textsuperscript{\dag}\\
\hspace{5mm} Cees G. M. Snoek\textsuperscript{*}%
\hspace{7mm} {Victor Rühle}\textsuperscript{\dag}%
\vspace{1.7mm}\\\textsuperscript{*}University of Amsterdam, \textsuperscript{\dag}Microsoft Research
}

\begin{document}
% \twocolumn[{%
% \renewcommand\twocolumn[1][]{#1}%
% \maketitle

% \vspace{-7.5mm}
% \begin{center}
%     \includegraphics[clip, width=\textwidth]{images/figure0.png}
%     \vspace{-1.1cm}
%     \captionof{figure}{Our model unlocks prompt comprehension, particularly spatial relationships between objects and attributing assignment. 
%     \cs{Change request: Let's follow the leading reading convention in science. Put the prompts on the left, and the methods on top, so you can compare from left to right per prompt and see that the best one is at the end (right column). Likely you need less prompts, and smaller images.}
%     }
%     \label{fig:teaser}
%     \vspace*{2.7mm}
% \end{center}%
% }]
\maketitle
\begin{abstract}
We propose \ours, an image generation pipeline that enhances spatial comprehension and attribute assignment in text-to-image generative models. Our pipeline enables the interpretation of instructions defining spatial relationships between objects in a scene, such as `An image of a gray cat on the left of an orange dog', and generate corresponding images.
This is especially important in order to provide more control to the user.
\ours{} overcomes the limitation of existing text-to-image diffusion models by decoding the generation of multiple objects into iterative steps: first generating a single object and then editing the image by placing additional objects in their designated positions. To create training data for spatial comprehension and attribute assignment we introduce a synthetic data generation process, that leverages a frozen large language model and a frozen layout-based diffusion model for object placement. 
We compare our approach to strong baselines and show that our model outperforms state-of-the-art image generation models in spatial comprehension and attribute assignment, despite being 3x to 5x smaller in parameters.
\end{abstract}
\vspace{-5mm}

\begin{figure}[h]
    \centering
    \includegraphics[trim={0cm 0cm 2cm 0cm},clip, width=\linewidth]{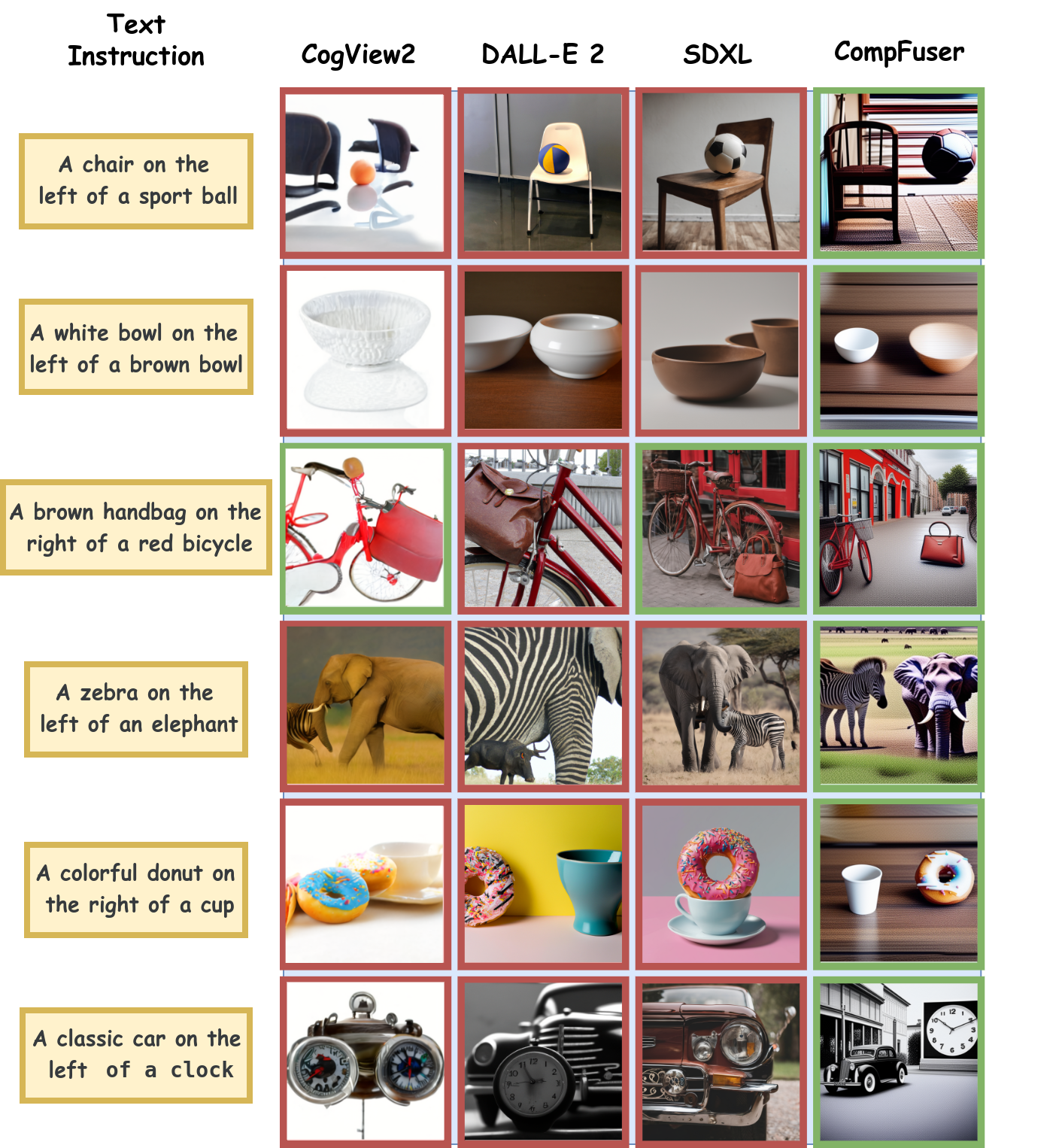}
    \caption{Examples from \ours. Different from existing text-to-image diffusion systems, we enable the understanding of prompts specifying spatial relationships between objects and their attribute assignment.}
    \label{fig:teaser}
    \vspace{-6mm}
\end{figure}
\section{Introduction}
\label{sec:intro}
% \begin{figure}[t]
%     \centering
%     % \includegraphics[width=0.98\linewidth]{images/figure_2_v2.png}
%     \includegraphics[width=\linewidth]{images/figure_2_v3.pdf}
%     \caption{The figure shows our \textit{\ours} model, which uses image instruction $c_I$ and text instruction $c_T$ to generate images, trained on our previously synthesized dataset.}
%     \label{fig:figure2}
% \end{figure}
% \begin{figure}[!t]
%     \centering
%     % First Figure
%     \begin{subfigure}[b]{\linewidth} % Adjust the width as needed
%         \includegraphics[width=\linewidth]{images/figure2_a.pdf} % Replace with your figure path
%         \caption{\textbf{Training pipeline.} The figure shows the training pipeline of our \textit{\ours} model, which is trained using image instruction $c_I$ and text instruction $c_T$ to generate images.}
%         \label{fig:figure2_a}
%     \end{subfigure}
    
%     % Second Figure
%     \begin{subfigure}[b]{\linewidth} % Adjust the width as needed
%         \includegraphics[trim={0.3cm 0 0 0}, clip, width=\textwidth]{images/figure2_b.pdf} % Replace with your figure path
%         \caption{\textbf{Inference pipeline.} The \textit{\ours} model operates as a dual-faceted generative mechanism, encompassing both text-and-image-to-image (T\&I2I) translation, depicted with a solid line, and text-to-image (T2I) synthesis, illustrated by a dashed line. Note that LDM refers to LLM-grounded diffusion \cite{li2023gligen}.}
%         \label{fig:figure2_b}
%     \end{subfigure}
%     \caption{\textbf{\ours} model.}
%     \label{fig:figure2}
% \end{figure}
\begin{figure*}[t]
    \centering
    \includegraphics[trim={0cm 0cm 0.5cm 0cm},clip, width=\linewidth]{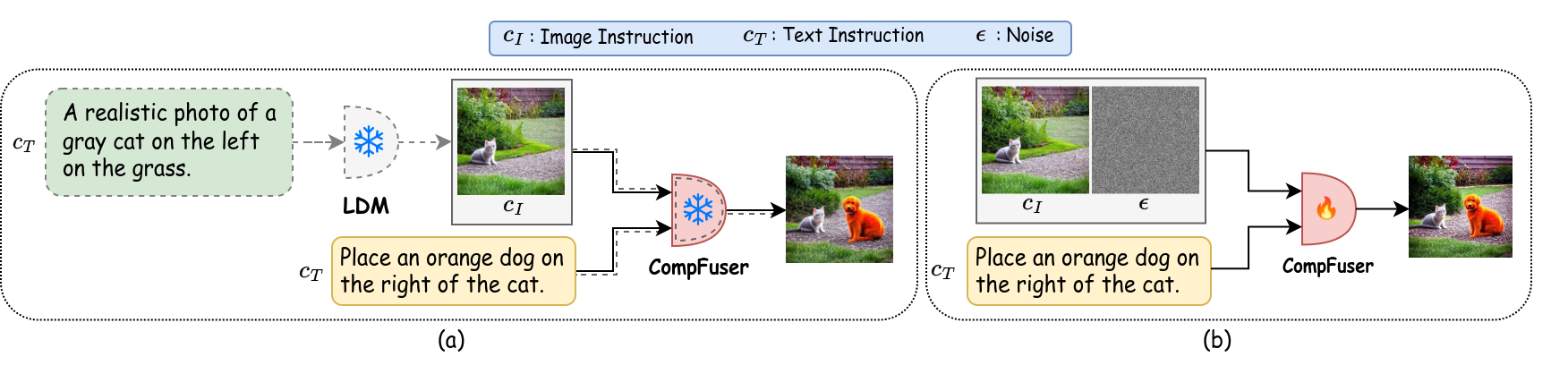}
    \vspace{-5mm}
    \caption{(a) \textbf{Inference pipeline.} \textit{\ours} employs an iterative generation process, first using text-to-image synthesis to generate a draft image (illustrated by dashed lines) and employs text-and-image-to-image generation to edit the image (depicted with solid lines). LDM refers to LLM-grounded diffusion \cite{li2023gligen}. (b) \textbf{Training pipeline.} This figure shows the training pipeline of our proposal, which is trained on synthesized data with image conditioning $c_I$ and text instruction $c_T$ to generate images.}
    \label{fig:figure2}
    \vspace{-5mm}
\end{figure*}

Text-to-image diffusion models have the potential to revolutionize digital image creation, by their ability to generate and edit artistic works and graphical designs \eg,~\cite{saharia2022photorealistic, ramesh2022hierarchical, Rombach_2022_CVPR, podell2023sdxl, Ho2020Denoising, peebles2023scalable}. Despite significant advancements, these models still face considerable challenges in understanding and executing complex prompts, especially regarding the spatial positioning among several objects and their attribute assignment \cite{gokhale2022benchmarking}. As illustrated in Fig.~\ref{fig:teaser}, this issue manifests itself in two primary aspects: (\emph{i}) current models exhibit a decline in performance when tasked with generating images containing two objects as opposed to their adeptness for single-object generation, and (\emph{ii}) even when these models successfully produce images featuring both objects, they often fail to adhere to the spatial arrangements detailed in the text instruction. This is typically because these models are either biased towards generating only the first object mentioned in the text prompt and ignoring the second one, showing better performance on commonly occurring object pairs, or having a tendency to merge two objects into one. This paper proposes a pipeline that enhances spatial comprehension and attribute assignment in text-to-image diffusion models, we call \ours.

% @@@Introduce InstructPix2Pix here, plus motivation.@@@

\textit{Attribute assignment} concerns the accurate linking of attributes to their respective entities, while \textit{spatial comprehension} involves understanding terms that describe objects' relative positioning. Others before us have explored the concept of spatial comprehension in text-to-image generative models \cite{nichol2022glide, Avrahami_2022_CVPR, zhang2023controllable, li2023gligen}. These studies primarily aim to enhance spatial comprehension through extra supervisor guidance such as user-generated masks and local CLIP-guided diffusion \cite{nichol2022glide, Avrahami_2022_CVPR}, incorporating extra inputs like bounding boxes and part keypoints \cite{li2023gligen}, or using segmentation masks \cite{zhang2023controllable}. However, they often overlook the fundamental aspect of comprehending text prompts alone. InstructPix2Pix by Brooks \etal~\cite{brooks2023instructpix2pix} is more aligned with our work, as it focuses on training a model to adhere to textual instructions for image editing. To do so,  InstructPix2Pix synthesizes a large dataset of image editing examples and trains a text-to-image diffusion model. This method shows promising results on modifying existing objects and backgrounds, but fails to follow edit instructions that requires spatial reasoning, such as the addition of new objects to an image. Inspired by InstructPix2Pix, we propose our \ours{} approach to improve spatial comprehension and attribute assignment.

% \cs{Here you start to talk to yourself and no longer to the reader, so it becomes very hard to follow where you want to go. Why do you need three paragraphs and a bullet summary to explain your method? I don't get it. Pix2Pix is used but never properly introduced. What are the contributions of this paper?}

% ------
% Others before us have explored the concept of spatial understanding in text-to-image generative models, as seen in works like \cite{nichol2022glide, Avrahami_2022_CVPR, zhang2023controllable, li2023gligen}. These studies primarily aimed to enhance spatial comprehension through techniques such as user-generated masks and local CLIP-guided diffusion \cite{nichol2022glide, Avrahami_2022_CVPR}, incorporating extra inputs like bounding boxes and part keypoints \cite{li2023gligen}, and using segmentation masks \cite{zhang2023controllable}. However, these methods often overlook the fundamental aspect of comprehending prompts, instead choosing to improve models with extra supervisory guidance. InstructPix2Pix by \cite{brooks2023instructpix2pix} is more aligned with our interest, as it focuses on training a model to adhere to textual instructions for image editing. Yet, this too has limitations in complex spatial reasoning tasks, especially when adding new objects to an image. Drawing inspiration from InstructPix2Pix, we introduce the \ours approach. 
% ----

We contribute in three major aspects. \textbf{First}, we propose an iterative generation process to enhance the capabilities of text-to-image generative models in terms of spatial comprehension and attribute assignment. We achieve this by leveraging an LLM to decode the original instruction into multiple generation steps, by first generating an image featuring a single object and then editing the image by placing additional objects in their specified positions, as shown in Fig. \ref{fig:figure2}  (a). This multi-step approach considerably enhances the spatial comprehension and attribute assignment capabilities of text-to-image diffusion models, leading to more contextually accurate and visually coherent image outputs.
\textbf{Second}, inserting an additional object into an existing scene using text instructions alone is not trivial. To accomplish this, we finetune an image editing model on a synthesized dataset for object placement. Specifically, we use an LLM to create detailed image layouts 
and use these layouts to guide the synthesis of image pairs using the LLM-grounded diffusion model \cite{lian2023llmgrounded}: one featuring a single object and the other showcasing both objects in their spatially defined positions. We then craft specific ``edit instructions'' based on the image pairs, explicitly detailing the spatial relationship between the objects. For example, ``Place xxx on the right of ooo.'' To enable spatial reasoning, we adapt InstructPix2Pix \cite{brooks2023instructpix2pix} on the synthetic images and edit instructions for object placement. \textbf{Third}, we empirically show that our approach outperforms seven text-to-image and text-and-image-to-image models in terms of spatial comprehension and attribute assignment. %, \cs{particularly in terms of $\text{VISOR}_{\text{uncond}}$, $\text{VISOR}_{\text{cond}}$, and Object Accuracy (OA) metrics.}

\section{Related work}
\label{sec:formatting}

\paragraph{Diffusion-based Text-to-Image Synthesis}

Diffusion models \cite{Sohl2015Deep, Song2019Generative, Ho2020Denoising} have emerged as a powerful family of generative models for image synthesis \cite{dhariwal2021diffusion, ho2022cascaded, saharia2023Image}. They are latent variable models characterized by a forward Markov chain process, where Gaussian noise is gradually added to the data. Diffusion models generate data by learning to reverse the forward process via an iterative process that reconstructs a denoised version of its input. Following the reparametrization in \cite{Ho2020Denoising}, the diffusion step trains a score network $\epsilon_\theta(x_t, t), t{=}1...T$ to predict the noise $\epsilon$ from $x_t$ to reconstruct $x_{t-1}$, where $\epsilon_\theta$ is often implemented with a UNet \cite{ronneberger2015u}. 

Diffusion models can be used for conditional generation by modeling a conditional distribution $p(x_t|c)$, where $c$ is the conditioning. This enables diffusion models to generate photo-realistic images from text prompts \cite{nichol2022glide, saharia2022photorealistic, ramesh2022hierarchical, Rombach_2022_CVPR}. 
Latent diffusion models, such as Stable Diffusion \cite{Rombach_2022_CVPR}, improves efficiency of diffusion models by using a variational autoencoder with encoder $\mathcal{E}$ and decoder $\mathcal{D}$ to map inputs from the original pixel space into a latent space $z{=}\mathcal{E}(x)$ and performs diffusion in the latent space instead of the pixel space. A conditional latent diffusion model is trained with the following denoising objective:
\begin{equation}
    L=\mathbb{E}_{\mathcal{E}(x), c, \epsilon \sim \mathcal{N}(0,1),t}\Bigl[ ||\epsilon - \epsilon_\theta(z_t, t, \tau_\theta(c))||^2\Bigr],
\end{equation}
where $\tau_\theta$ is a domain-specific encoder that projects the conditioning $c$ to an intermediate representation. For conditional text-to-image generation \cite{saharia2022photorealistic, nichol2022glide, ramesh2022hierarchical}, the conditioning is often projected with a pretrained text encoder, such as CLIP \cite{radford2021learning}. CLIP employs a contrastive objective to jointly train an image encoder and a text encoder to predict the correct pairings of image and caption text, which allows it to align the text and image domains. $\epsilon_\theta(z_t, c)$ is a conditional score network with classifier-free guidance \cite{ho2022classifier}, which is realized by augmenting the UNet backbone with the cross-attention mechanism \cite{vaswani2017attention} in Stable Diffusion \cite{Rombach_2022_CVPR}.

\paragraph{Diffusion Models for Image Editing}
Diffusion models have also been applied for text-guided image editing. For example, GLIDE \cite{nichol2022glide} can edit image by applying a user-drawn mask around areas of interest in the image and employing CLIP-guided inpainting to edit the masked area. Similarly, Blended Diffusion \cite{Avrahami_2022_CVPR} uses local CLIP-guided diffusion in the masked area for region-based edits. Recent works have extended image editing to using text prompt alone for editing instructions, eliminating the need to apply a manual mask. Prompt-to-prompt \cite{hertz2022prompt} allows for generating images from edited prompts, such as replacing existing tokens with new entities or adding descriptive tokens to the original prompt, by fixing the attention maps of the preserved tokens. DiffEdit \cite{couairon2023diffedit}, Null-text Inversion \cite{mokady2023null} and DDPM Inversion \cite{HubermanSpiegelglas2023} edits an image by employing DDIM \cite{song2021denoising} or DDPM \cite{Ho2020Denoising} inversion to generate an reverse diffusion trajectory and regenerates the image with a modified prompt. DreamBooth \cite{ruiz2023dreambooth} allows for the preservation of an object in image editing by finetuning a model to learn the representation of that object from a small set of images. 
Our work is primarily inspired by InstructPix2Pix \cite{brooks2023instructpix2pix}, which leverages Prompt-to-Prompt to generate synthetic data for finetuning a model to follow explicit edit instructions. However, previous research on image editing has mainly focused on modifying existing objects and backgrounds and they often fail to follow edit instructions that requires spatial reasoning, such as the addition of new objects to the image.

\section{Methods}
\begin{figure*}[t]
    \centering
    \includegraphics[width=\textwidth]{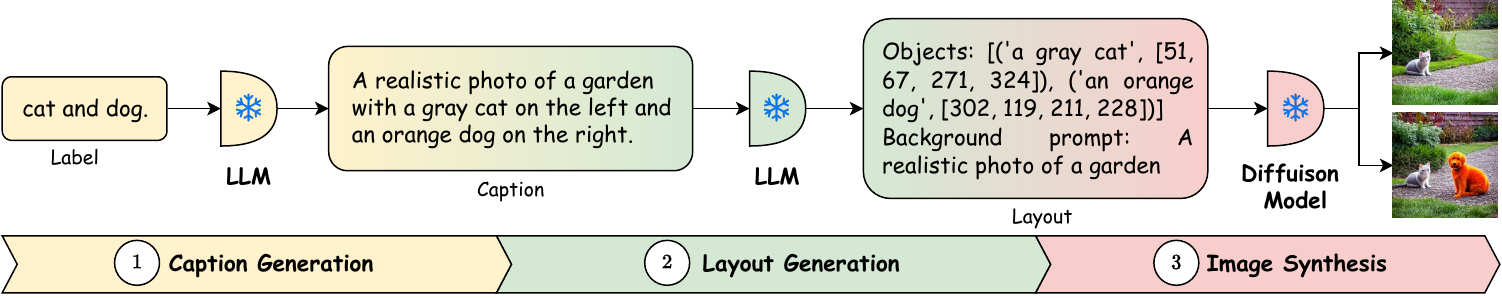}
    \caption{\textbf{Data synthesis.} Step (1): Initiating the dataset synthesis, we first generate captions using LLMs that simulate attribute assignment and spatial reasoning scenarios. Step (2): Leveraging LLMs, we create an image layout based on the caption from Step (1), which encompasses two core elements: instance-level annotations delineated by a set of captioned bounding boxes, and a background prompt. Step (3): We employ the LLM-grounded diffusion model~\cite{lian2023llmgrounded} to generate images based on the layout defined in Step (2), ensuring a singular object distinction between the two resultant images.}
    \label{fig:figure1}
    \vspace{-5mm}
\end{figure*}

% \begin{figure*}[h]
%     \centering
%     \includegraphics[width=0.98\textwidth]{images/figure1.png}
%     \caption{Initiating the dataset synthesis, we first generate captions using LLMs that simulate attribute assignment and spatial reasoning scenarios (a). Leveraging LLMs, we create an image layout based on the caption from Step (a), which encompasses two core elements: instance-level annotations delineated by a set of captioned bounding boxes, and a background prompt (b). In Step (c), we employ the LLM-grounded Diffusion model~\cite{lian2023llmgrounded} to generate images based on the layout defined in Step (b), ensuring a singular object distinction between the two resultant images.}
%     \label{fig:figure1}
% \end{figure*}

% \begin{figure}[h]
%     \centering
%     \includegraphics[width=\linewidth]{images/figure1_v2.png}
%     \caption{Initiating the dataset synthesis, we first generate captions using LLMs that simulate attribute assignment and spatial reasoning scenarios (a). Leveraging LLMs, we create an image layout based on the caption from Step (a), which encompasses two core elements: instance-level annotations delineated by a set of captioned bounding boxes, and a background prompt (b). In Step (c), we employ the LLM-grounded Diffusion model~\cite{lian2023llmgrounded} to generate images based on the layout defined in Step (b), ensuring a singular object distinction between the two resultant images.}
%     \label{fig:figure1}
% \end{figure}
\label{sec:methods}
Our objective is to generate images that can adhere to complex spatial relational instructions. To achieve this, we propose an iterative generation pipeline \ours. The pipeline is capable of editing existing images by placing additional objects at specified positions (Fig. \ref{fig:figure2}). To enable spatial reasoning for diffusion models, we synthesize a multimodal dataset by leveraging pre-trained LLMs in tandem with LLM-grounded diffusion models \cite{lian2023llmgrounded}. We adapt InstructPix2Pix \cite{brooks2023instructpix2pix} on the synthesized dataset for adding an object to an existing scene. In the subsequent sections, we provide some background on InstructPix2Pix and LLM-grounded Diffusion (LMD) (Sec. \ref{subsec:preliminaries}) and present the details of \ours, including dataset synthesis for spatial reasoning (Sec. \ref{subsec:dataset_synthesis}) as well as training and inference pipelines of \ours{} (Sec. \ref{subsec:instobj2scene}).

\subsection{Preliminaries}

\label{subsec:preliminaries}
\paragraph{In-depth Look at InstructPix2Pix:} InstructPix2Pix \cite{brooks2023instructpix2pix} is an image editing model based on conditional latent diffusion. InstructPix2Pix employs Prompt-to-Prompt \cite{hertz2022prompt} to generate synthetic training data and trains a latent diffusion model conditioned on an image and a text instruction prompt to generate a modified variant of the image. The score network $\epsilon_\theta(z_t, c_I, c_T)$ has both image $c_I$ and text instruction $c_T$ conditionings. To leverage classifier-free guidance for both conditionings with guidance scales $\omega_I$ and $\omega_T$, the modified score estimate is:
\begin{align}
    \label{eq:instruct_pix2pix}
    \tilde{\epsilon}_\theta(z_t, c_I, c_T) &= \epsilon_\theta(z_t, \varnothing, \varnothing) \nonumber \\ 
    &+ \omega_I\bigl(\epsilon_\theta(z_t, c_I, \varnothing) - \epsilon_\theta(z_t,\varnothing, \varnothing)\bigr) \nonumber \\
    &+\omega_T\bigl(\epsilon_\theta(z_t, c_I, c_T) - \epsilon_\theta(z_t, c_I, \varnothing)\bigr).
\end{align}
The learning objective for image editing is:
\begin{equation}
\label{eq:controlnet}
\resizebox{.9\linewidth}{!}{$L=\mathbb{E}_{\mathcal{E}(x), c_I, c_T, \epsilon \sim \mathcal{N}(0,1),t}\Bigl[ ||\epsilon - \epsilon_\theta(z_t, t, c_I, \tau_\theta(c_T))||^2\Bigr].$}
\end{equation}

\paragraph{LLM-grounded Diffusion:} Our data generation pipeline is inspired by LLM-grounded Diffusion (LMD) \cite{lian2023llmgrounded}, which augments text-to-image generation with a large language model. LMD uses an LLM to produce an image layout from the text description, which consists of a background and bounding boxes for foreground objects. It then applies layout-guided stable diffusion to generate the image. To do that, it first generates masked latent inversions $z_t^{(i)}$ for each bounding box object $i$ with DDIM Inversion \cite{song2021denoising} and then creates a composed latent $z_t^{(comp)}$ by combining the masked latent inversions with a background latent $z_t^{(b)}$ initialized with a Gaussian noise. 
\begin{equation}
\resizebox{.9\linewidth}{!}{$z_t^{(comp)} = z_t^{(b)} \cdot (1-m^{(i)}) + z_t^{(i)} \cdot m^{(i)}, \forall i, t = T,...,T-k,$}
\end{equation}
where $m^{(i)}$ is the mask for the foreground object $i$ and $\cdot$ is element-wise multiplication. $k\in[0,T]$ is a hyperparameter for the number of timesteps to apply the composition, which balances object control and overall image coherency.

When generating the masked latent inversion $z_t^{(i)}$ of each foreground object, it introduces an attention control energy function $E(\cdot)$ before each denoising step to strengthen the cross-attention between the pixels inside the bounding box with its corresponding object caption:
\begin{equation}
\begin{aligned}
    E(\mathbf{A}^{(i)}, i, v) &= -\text{Topk}_u(\mathbf{A}_{uv} \cdot m^{(i)}) \\
    &+ \omega \text{Topk}_u(\mathbf{A}_{uv} \cdot (1-m^{(i)})).
\end{aligned}
\end{equation}
\begin{equation}
    z_{t}^{(i)} \gets z_{t}^{(i)} - \eta\nabla_{z_{t}^{(i)}}\sum_{v \in V} E(\mathbf{A}^{(i)}, i, v)
\end{equation}
where $\mathbf{A}_{uv}=\text{Softmax}(\mathbf{q}_u^\top\mathbf{k}_v)$ and $\mathbf{q}_u$ and $\mathbf{k}_v$ are linear transformations of the image feature at spatial location $u$ and text feature at token index $v$ in the prompt, $V$ contains all the token indices for the current object box caption, $\text{Topk}_u$ takes the average of top-k values across the spatial dimension $u$, and $\omega$ and $\eta$ are strength parameters.

\subsection{Dataset Synthesis}
\label{subsec:dataset_synthesis}
We curate a dataset for a distinctive image editing task, where our goal is to train an image editing model to interpret and reason about spatial relations between objects and accurately position an object relative to an existing object in the image. Our dataset synthesis is structured in a three-stage generation process, as illustrated in Fig. \ref{fig:figure1}. In the subsequent sections, we provide a detailed explanation of each stage.

\subsubsection{Caption Generation} 
\label{subsec:caption_generation}
We initiate our dataset synthesis process by generating captions that imitate the spatial reasoning scenarios. Specifically, we focus on captions with two objects, one positioned on the left (\textcolor{bargreen}{green}) and the other on the right (\textcolor{barblue}{blue}), each preceded by one adjective, such as colors, shapes, and style. 
Each caption also begins with a background text (\textcolor{barpurpule}{purple}) to set the image's context. See following for the desired caption format: 
\begin{tcolorbox}[colback=red!5!white,colframe=red!75!black, width=.48\textwidth, size=small,title=Caption Format,]
\noindent\textcolor{barpurpule}{A realistic photo of a garden} with \textcolor{bargreen}{a gray cat on the left} and \textcolor{barblue}{an orange dog on the right}.
\end{tcolorbox}
To achieve this, we harness the capabilities of LLMs. Given its extensive training on diverse internet data, an LLM effectively crafts complex captions tailored to specific scenarios. We explicitly provide the language model with a detailed task description as a prompt, which is crafted to direct the model's behavior toward generating the desired captions that include the two objects specified within the prompt. Moreover, LLMs can also provide more realistic context to the original object categories for generating the image, e.g. by determining the \textit{background prompt}. 
% \mx{We could also mention how we expanded the diversity of the prompt here.}
% \begin{tcolorbox}[colback=red!5!white,colframe=red!75!black, width=.48\textwidth, size=small,title=Caption Generation Prompt]
% \noindent \textbf{[Task Description]:} You are an intelligent caption generator for a photo, image, or painting. The caption must contain two primary objects. I will provide the objects's name. Each object is better to be described with one adjective. Keep the caption short, and do not be verbose. Generate the background context accordingly. 
% \newline
% \newline
% \textbf{[Objects]:} cat and dog.
% \newline
% \textbf{[Caption]:} $<\text{LLM-generated}>$
% \end{tcolorbox}
%
However, LLMs have a tendency to produce unintended objects categories that may not coexist in a single image, e.g.  \textit{a spacecraft} and \textit{a shark}. To address this issue, we utilize the MSCOCO object detection dataset \cite{lin2014microsoft} to construct a co-occurrence matrix for object categories. This matrix quantifies the frequency at which object category pairs co-occur within the same images in the dataset. Leveraging this matrix, we sample pairs of commonly co-occurring objects and input them into the language model to generate captions as illustrated in Fig. \ref{fig:figure1} part 1. Moreover, we also supply the language model with in-context learning demonstrations subsequent to the task description. These demonstrations clarify our captioning intent, ensuring the desired caption format and reducing potential hallucination.
\begin{tcolorbox}[colback=red!5!white,colframe=red!75!black, width=.48\textwidth, size=small,title=Caption Generation Prompt with ICL]
\noindent \textbf{[Task Description]:} You are an intelligent caption generator for a photo, image, or painting. The caption must contain two primary objects. I will provide the objects's name. Each object is better to be described with one adjective. Keep the caption short and do not be verbose. Generate the background context accordingly.
\newline
$\cdot\cdot\cdot$
\newline
\textbf{[Objects]:} teddy bear and book.
\newline
\textbf{[Caption]:} A realistic photo of a wooden table with a book on the right and a teddy bear on the left.
\newline
$\cdot\cdot\cdot$
\newline
\textbf{[Objects]:} cat and dog.
\newline
\textbf{[Caption]:} $<\text{LLM-generated}>$
\end{tcolorbox}

\subsubsection{Layout Generation}
\label{subsec:layout_generation}
In this section, we utilize LLMs to produce an image layout based on the caption generated in the section \ref{subsec:caption_generation}, illustrated in Fig. \ref{fig:figure1} part 2. Each constructed layout consists of two primary components: instance-level annotations using a collection of captioned bounding boxes (\textcolor{bargreen}{green} and \textcolor{barblue}{blue}) and a background prompt (\textcolor{barpurpule}{purple}). See following for the desired layout format:
\begin{tcolorbox}[colback=red!5!white,colframe=red!75!black, width=.48\textwidth, size=small,title=Layout Format,]
\noindent Objects: [\textcolor{bargreen}{('a gray cat', [51, 67, 271, 324])}, \textcolor{barblue}{('an orange dog', [302, 119, 211, 228])}]
\newline
\textcolor{barpurpule}{Background prompt: A realistic photo of a garden with}
\end{tcolorbox}
Instance-level annotations are defined as \texttt{(caption,  [x, y, w, h])}, where \texttt{[x, y, w, h]} specifies the bounding box location of each object and \texttt{caption} determines the content of each bounding box. However, we found that decomposing foreground objects from background is challenging due to the hallucination of LLMs. To address this, similar to previous step, we stick to the in-context learning capabilities of LLMs and precisely control the layout generation through two key designs. First, we ensure that every object instance is associated to a single bounding box. Second, any foreground objects detailed within the boxes should not exist in the background prompt. This ensures that our layout-grounded image generator exclusively manages all foreground objects. See appendix for layout generation prompt with ICL.

\subsubsection{Image Synthesis}
\label{subsec:image_synthesis}
In this section, our focus shifts to synthesizing images based on the layouts generated in the section \ref{subsec:layout_generation}, illustrated in Fig. \ref{fig:figure1} part 3. For demonstrative purposes, consider a primary layout comprising two objects, as shown below:
\begin{tcolorbox}[colback=red!5!white,colframe=red!75!black, width=.48\textwidth, size=small,title=$\mathcal{L}_2$: Two-objects Layout,]
\noindent 
Objects: [($\text{caption}_A$, $\text{bbox}_A$), ($\text{caption}_B$, $\text{bbox}_B$)]\newline
Background prompt: $<\text{background text}>$
\end{tcolorbox}
\noindent In this layout, each object, denoted as $A$ and $B$, is annotated individually. We then generate a variant layout $\mathcal{L}_1$ by selectively removing the annotation of one object (e.g., object $A$), keeping the rest unchanged. In the example below, the $A$ object's layout information is excluded, leaving only the $B$ object's layout:

\begin{tcolorbox}[colback=red!5!white,colframe=red!75!black, width=.48\textwidth, size=small,title=$\mathcal{L}_1$: One-object Layout,]
\noindent
Objects: [($\text{caption}_B$, $\text{bbox}_B$)]\newline
Background prompt: $<\text{background text}>$
\end{tcolorbox}

\noindent Utilizing these layouts, we employ the LLM-grounded Diffusion model~\cite{lian2023llmgrounded} to generate image instruction $c_I$ and the input image $x$ defined in Eq \ref{eq:instruct_pix2pix}. Here, $c_I$ corresponds to the one-object layout $\mathcal{L}_1$ while $x$ represents the two-objects layout $\mathcal{L}_2$. During the image generation, we keep all hyper-parameters such background and foreground seeds same to ensure uniformity in perspective, style, and lighting across the produced images. 

\subsubsection{Background Retention}
\begin{figure}[h]
    \centering
    \includegraphics[width=\linewidth]{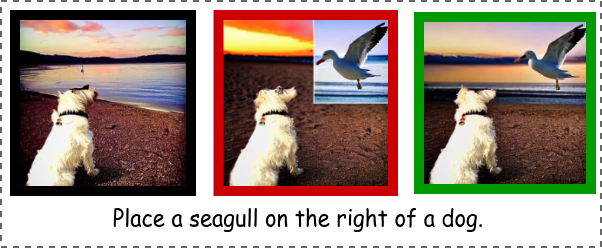}
    \caption{\textbf{Example of Background Retention:} With the original LDM method, inserting an object into an image (black) alters the background and creates incoherent blocks around the added object (red). Our background retention method (green) effectively preserves background details of input image.}
    \label{fig:retention_motivation}
    \vspace{-2mm}
\end{figure}
\label{subsec:background_retention}
During the image synthesis process, we detect subtle yet noticeable changes in the background details of images $c_I$ and $x$ (Fig. \ref{fig:retention_motivation}). These changes seem to adversely impact the efficacy of our generative model. To mitigate this issue, we propose a novel energy function aimed at preserving the background details. This function works by transferring the background details from image $c_I$ to image $x$. Let $\mathbf{A}^{(1)}$ and $\mathbf{A}^{(2)}$ represent the reference and target cross-attention maps within the score network $\epsilon_\theta$, corresponding to images $c_I$ and $x$, respectively. These maps are defined as follows:
\begin{align}
    \mathbf{A}^{(i)}_{uv} = \text{Softmax}(\mathbf{q}^\top_u \mathbf{k}_v)
\end{align}
where $\mathbf{q}^\top_u$ denotes the cross-attention query at spatial location $u$, and $\mathbf{k}_v$ represents the key corresponding to the text feature at token index $v$ in the text prompt. 
Let $V$ be a set of token indices in the text prompt associated to the background prompt and the caption of first object. Let also $m_2$ be a binary mask of the second object with the region in the bounding box set 1. We preserve the background detail of first image and the second image by transferring the reference cross-attention $\mathbf{A}^{(1)}$ to target cross-attention $\mathbf{A}^{(2)}$. To achieve this, we define an energy function:
\begin{align}
\resizebox{.9\linewidth}{!}{$E(\mathbf{A}^{(1)}, \mathbf{A}^{(2)}) = \frac{1}{2}\norm{(1-m_2) \cdot \left[\sum_{u,v \in V}{\mathbf{A}^{(1)}_{uv}} - \sum_{u,v \in V}{\mathbf{A}^{(2)}_{uv}}\right]}^2$}
\end{align}
This energy function is minimized by updating the latent $z$ during the diffusion process. This approach aims to align the target cross-attention with the reference cross-attention signal for identical tokens $V$ and pixels associated with both the background region and the first object.

% \mx{It was a bit hard to understand what background retention does just from the description. I was thinking that maybe we can move figure 6 forward to this section and refer to it. But it would also make sense to have figure 6 later in the results. What do you think?} \mo{I have also the same feeling. Probably need to add another figure to the text explaining this loss function? good idea}

\subsubsection{Edit Instruction Generation}
\label{subsec:Generating_text_instruction}
In this section, we construct an edit instruction that specifies the spatial relationship between two objects $A$ and $B$ with respect to layouts $\mathcal{L}_1$ and $\mathcal{L}_2$. Let $\mathcal{R}$ be the set of spatial relationships between objects. In this paper, we only focus on two-dimensional relationships $\mathcal{R}=\{left, right\}$. Let $\text{bbox}_{A} = [x_{A}, y_{A}, w_A, h_A]$ and $\text{bbox}_{B} = [x_{B}, y_{B}, w_B, h_B]$ be the spatial location of objects $A$ and $B$, respectively. The relationship between object $B$ with respect to $A$ is determined by:
\begin{align}
R_{B \rightarrow A} = \begin{cases} 
      \text{left}   & \texttt{if } \text{cnt}_{x,B}   < \text{cnt}_{x,A} \\
      \text{right}  & \texttt{elif } \text{cnt}_{x,B} > \text{cnt}_{x,A} \\
      \text{n/a}    & \texttt{otherwise}
   \end{cases}
\end{align}
, where $\text{cnt}_{A}$ and $\text{cnt}_{B}$ are the center of each object bounding boxe computed using following formulation:
\begin{equation}
\text{cnt}_{A} = (\text{cnt}_{x,A}, \text{cnt}_{y,A}) = (x_A + (\frac{w_A}{2}), y_A+(\frac{h_A}{2}))
\end{equation}
\begin{equation}
\text{cnt}_{B} = (\text{cnt}_{x,B}, \text{cnt}_{y,B}) = (x_B + (\frac{w_B}{2}), y_B+(\frac{h_B}{2}))
\end{equation}
Given the caption related to objects $A$ and $B$ and their relationship $R_{B \rightarrow A}$, we construct the text instruction $c_{T}$ using the following format:
\begin{tcolorbox}[colback=red!5!white,colframe=red!75!black, width=.48\textwidth, size=small,title=Edit Instruction Format]
\noindent Place $[\text{caption}_B]$ on the $[R_{B \rightarrow A}]$ of $[\text{caption}_A]$.
\end{tcolorbox}
\subsection{\ours}
\label{subsec:instobj2scene}
\paragraph{Training Pipeline.} Let $X = \{(c_I^i, c_T^i, x^i)\}_{i=1}^{N}$ be the set of training data containing synthesized image conditioning $c_I^i$, text instruction $c_T^i$ and output image $x^i$ that is to be predicted used in Eq. \ref{eq:instruct_pix2pix}. Given $X$, we finetune an image editing model to place the object, as illustrated in  Fig. \ref{fig:figure2} (b). Notebly, instead of initializing it randomly, we use a pretrained version of InsturctPix2Pix \cite{brooks2023instructpix2pix} and adapt its parameters efficiently using the parameter efficient fine-tuning such LoRA \cite{hu2021lora}. 

\paragraph{Inference Pipeline.} During the inference phase, the \ours model functions as a text-to-image model. Specifically, this model initially creates the scene's first object based on a text instruction, represented by dashed lines in Fig. \ref{fig:figure2} (a). Following the creation of this initial object as an image conditioning $c_I$, our \ours editing technique is employed to integrate the second object into the existing scene given a textual instruction $c_T$. This process is defined as a text-and-image-to-image  generative model, visually represented by solid lines in Fig. \ref{fig:figure2} (a). 

To improve the image quality produced by \ours, we integrate the Stable Diffusion XL (SDXL) model with the DDIM method, as outlined in \cite{song2020denoising}. Our approach involves initially introducing noise into the image via DDIM inversion during diffusion process. Next, we use the DDIM process and the SDXL-base model together in reverse diffusion to eliminate this noise and restore the image. The final step involves enhancing the image quality using the SDXL-enhancer model. Each phase is designed to incrementally refine the image, balancing the correctness of the relationship between objects and efficiency in the enhancement process by leveraging pretrained SDXL models (See appendix for more detail).

\section{Experimental setup}
\label{sec:exper_setup}

\paragraph{Implementation details.}
For caption and layout generation, we use gpt-3.5-turbo \cite{brown2020language}, leveraging its in-context learning abilities for generating precise captions and layouts. GLIGEN \cite{li2023gligen} is employed for the generation of input-output image pairs used for the adaptation stage. Our \ours{} is developed using PyTorch and Diffusers \cite{von2022diffusers}, and we plan to make the implementation publicly available for further research. Details about the implementation and hyperparameters can be found in the appendix.

\paragraph{Datasets.} 
For the adaptation and evaluation of \ours, we propose a text-and-image-to-image dataset that includes distinct train and test splits tailored for spatial comprehension and attribute assignment tasks. 
%Specifically, spatial reasoning involves understanding terms that describe objects' relative positions, like \textit{left} or \textit{right}. 
We craft $22{,}000$ instruction triplets for this purpose: $20{,}000$ for fine-tuning and $2{,}000$ as a held-out set for evaluation. These splits will be made available to encourage further research.

\paragraph{Baselines.}
Following \cite{gokhale2022benchmarking}, we compare our \ours{} model with various baselines. The models we evaluate against include GLIDE \cite{nichol2022glide}, Stable Diffusion 1.5 (SD1.5) \cite{Rombach_2022_CVPR}, Stable Diffusion XL (SDXL) \cite{podell2023sdxl}, InstructPix2Pix \cite{brooks2023instructpix2pix}, DALL-E mini \cite{Dayma_DALL·E_Mini_2021}, CogView2 \cite{ding2022cogview2}, and different variants of Compositional Diffusion Models including GLIDE+CDM, SD1.5+CDM, SDXL+CDM \cite{liu2022compositional}. We analyze these models based on their input types, as detailed in Tables \ref{tab:t-and-i-2i} and \ref{tab:t2i}. Models like GLIDE, Compositional Diffusion Model variants, DALL-E mini, and CogView2 are text-to-image models, using only text instructions. In contrast, Stable Diffusion 1.5, Stable Diffusion XL, and \ours are both text-to-image and text-and-image-to-image models, capable of handling text, image, or both inputs as conditioning signal. InstructPix2Pix is a text-and-image-to-image diffusion model, requiring both text instruction and image conditioning for image generation.

\paragraph{Training details.}
Our \ours{} is trained using mixed-precision with Bfloat16~\citep{abadi2016tensorflow}. In the adaptation stage, we use a batch size of $8$ over $12{,}000$ iterations and 4 A5000 GPUs. Furthermore, we use the AdamW optimizer~\citep{kingma2014adam} with a learning rate of $1e$-$4$ and a linear learning rate schedular. Our \ours{} is initialized using the pre-trained weights of InstructPix2Pix \cite{brooks2023instructpix2pix}, chosen for its superior performance in global edits, such as texture and lighting adjustments. We fine-tune our model using LoRA while keeping the pre-trained weights fixed.

\paragraph{Evaluation criteria.}
Following \cite{gokhale2022benchmarking}, we assess our method by employing the readily available open vocabulary object detector, OWL-ViT \cite{minderer2022simple}, to extract bounding boxes for target objects. We subsequently verify if each produced image aligns with the given prompt in terms of layout definition. Formally,  Let $\mathcal{V}$ be any pretrained object detector that returns a set of detected objects in image $x$ from
a set of categories $C$. Then, object accuracy for an image x, generated by a
text instruction containing objects $A$ and $B$ is:
\begin{align}
    \text{OA}_{x}(A, B) = \mathbbm{1}_{\mathcal{V}(x)}(A \cap B) 
\end{align}
Object accuracy is agnostic to the spatial relationship between objects categories $A$ and $B$. To capture spatial relationship in our generation pipeline, we harness VISOR metric introduced in \cite{gokhale2022benchmarking}. 
\begin{align}
\resizebox{0.88\linewidth}{!}{
$S = \sum_{i=1}^{N} s_i, \quad
s_i = \begin{cases} 
      1 & \text{if } \text{OA}_x = 1 \text{ and } R_{\text{gen}} = R \\
      0 & \text{otherwise}
   \end{cases}$
}
\end{align}
\begin{align}
\resizebox{0.88\linewidth}{!}{
$U = \sum_{i=1}^{N} u_i, \quad
u_i = \begin{cases} 
      1 & \text{if } \text{OA}_x = 1 \text{ and } R_{\text{gen}} \neq R \\
      0 & \text{otherwise}
   \end{cases}$
}
\end{align}
\begin{align}
    \text{VISOR}_{\text{uncond}} &= S/N, & 
    \text{VISOR}_{\text{cond}} &= S/(S + U)
\end{align}
, where $R_{\text{gen}}$ is spatial
relationship between the objects in the generated image and $R$ is the ground-truth relationship between the objects provided in the text instruction. 
$\text{VISOR}_{\text{uncond}}$ computes the joint probability of both accurate generation and correct relative positioning of a pair of objects in a scene.
$\text{VISOR}_{\text{cond}}$ is defined as the conditional likelihood of correct spatial relationships being generated, given that both objects were generated correctly.

\section{Evaluation}
\label{sec:evaluation}

\subsection{Quantitative Evaluation}
\label{sec:quantitative_evaluation}
In this section, we assess our \ours{} framework by benchmarking it against established baselines: text-to-image and text-and-image-to-image models.

\begin{table}
\centering
\resizebox{1\linewidth}{!}{%
\begin{tabular}{lccccc}
\toprule
 &  & &  & \multicolumn{2}{c}{\textbf{VISOR (\%)}} \\
 \cmidrule(lr){5-6}
\textbf{Model} & $c_T$ & $c_I$ & \textbf{OA (\%)} & \textbf{uncond} & \textbf{cond} \\
\midrule
SD1.5 \cite{Rombach_2022_CVPR} & \cmark & \cmark & \underline{30.81} & \underline{23.17} & 75.24 \\
\midrule
SDXL \cite{podell2023sdxl} & \cmark & \cmark & 12.25 & 10.17 & \underline{80.21} \\
\midrule
InstructPix2Pix \cite{brooks2023instructpix2pix} & \cmark & \cmark & 26.18 & 20.23 & 77.70 \\
\midrule
\rowcolor{palered}\textbf{\ours} & \cmark & \cmark & \textbf{57.32} & \textbf{51.63} & \textbf{90.73} \\
\bottomrule
\end{tabular}
}
\caption{This figure compares various text-and-image-to-image models, assessing them based on Object Accuracy (OA) and each iteration of VISOR. Our proposed model notably exceeds the performance of InstructPix2Pix, achieving a substantially higher accuracy. Note that $c_T$ and $c_I$ represents text and image conditionings, respectively.}
\label{tab:t-and-i-2i}
\end{table}
\paragraph{Comparison with text-and-image-to-image baselines.} Table \ref{tab:t-and-i-2i} compares our \ours{} model with other text-and-image-to-image models. The results show that our model excels at incorporating both text and image instructions to add a second object into an image, achieving $27\%$ higher object accuracy. For the $\text{VISOR}_{\text{uncond}}$ and $\text{VISOR}_{\text{cond}}$ metrics, our model does better than the others by $27\%$ and $10\%$, respectively. This highlights that our model is not just good at getting both objects into the image, but also at placing them correctly in relation to each other. Notably, \ours{} surpasses InstructPix2Pix in object accuracy, $\text{VISOR}_{\text{uncond}}$, and $\text{VISOR}_{\text{cond}}$ by $31.14\%$, $31.40\%$, and $13.03\%$ respectively. This performance highlights the effectiveness of our synthesized dataset in improving the spatial comprehension of InstructPix2Pix, upon which our proposed model is built.

\begin{table}
\centering
\resizebox{1\linewidth}{!}{%
\begin{tabular}{lccccc}
\toprule
 &  &  &  & \multicolumn{2}{c}{\textbf{VISOR (\%)}} \\
 \cmidrule(lr){5-6}
\textbf{Model} & $c_T$ & $c_I$ & \textbf{OA (\%)} & \textbf{uncond} & \textbf{cond} \\
\midrule
GLIDE \cite{nichol2022glide} & \cmark & \xmark & 11.02 & 05.51 & 50.00 \\
+ CDM \cite{liu2022compositional} & \cmark & \xmark& 09.44 & 04.72 & 50.00 \\
\midrule
DALLE-mini \cite{Dayma_DALL·E_Mini_2021} & \cmark & \xmark& 26.63 & 12.96 & 48.69 \\
\midrule
CogView2 \cite{ding2022cogview2} & \cmark & \xmark& 21.95 & 10.51 & 47.89 \\
\midrule
SD1.5 \cite{Rombach_2022_CVPR} & \cmark & \xmark& 34.59 & 16.81 & 48.59 \\
+ CDM \cite{liu2022compositional} & \cmark & \xmark & 35.06 & 16.53 & 47.17 \\
\midrule
SDXL \cite{podell2023sdxl} & \cmark & \xmark & \textbf{59.51} & \underline{29.45} & \underline{49.49} \\
+ CDM \cite{liu2022compositional} & \cmark & \xmark & 41.31 & 20.05 & 48.54 \\
\midrule
\rowcolor{palered}\textbf{\ours} & \cmark & \xmark &  \underline{57.32} & \textbf{51.63} & \textbf{90.73} \\
\bottomrule
\end{tabular}
}
\caption{This figure compares various text-to-image models, focusing on Object Accuracy (OA) and different iterations of VISOR. Notably, our proposed model substantially surpasses the performance of SDXL, demonstrating a significant improvement. Note that $c_T$ and $c_I$ represents text and image conditionings, respectively.}
\label{tab:t2i}
\vspace{-5mm}
\end{table}

\paragraph{Comparison with text-to-image baselines.} The comparative results against text-to-image models, outlined in Table \ref{tab:t2i}, indicate a disparity in object accuracy; all models, with the exception of \ours{} and SDXL, exhibit accuracy below $40\%$. This suggests their inability to consistently generate images with both objects referenced in the input prompt. Notably, \ours{} and SDXL perform similarly and demonstrate superior performance, achieving more than a $16\%$ increase in object accuracy over their counterparts. Regarding the $\text{VISOR}_{\text{uncond}}$ metric, our \ours{} model significantly surpasses all baseline models, with a substantial margin of $20\%$. This performance underlines our model's enhanced capability to accurately generate images that not only include both target objects but also depict their interrelations within the image correctly. 
The advantage of our \ours{} model becomes even more pronounced when considering the $\text{VISOR}_{\text{cond}}$ metric. Here, our framework surpasses competing models by a substantial $40\%$ margin in cases where both objects are accurately generated. Moreover, our model with 1 billion parameters surpasses the spatial comprehension and attribute assignment of larger counterparts, e.g. Stable Diffusion XL (3.5 billion parameters) and Glide (5 billion parameters) to name a few, in performance.

\subsection{Qualitative Evaluation}
\label{sec:qualitative_evaluation}
\begin{figure}[t]
    \centering
    \includegraphics[trim={0cm 0cm 0cm 0cm},clip,width=\linewidth]{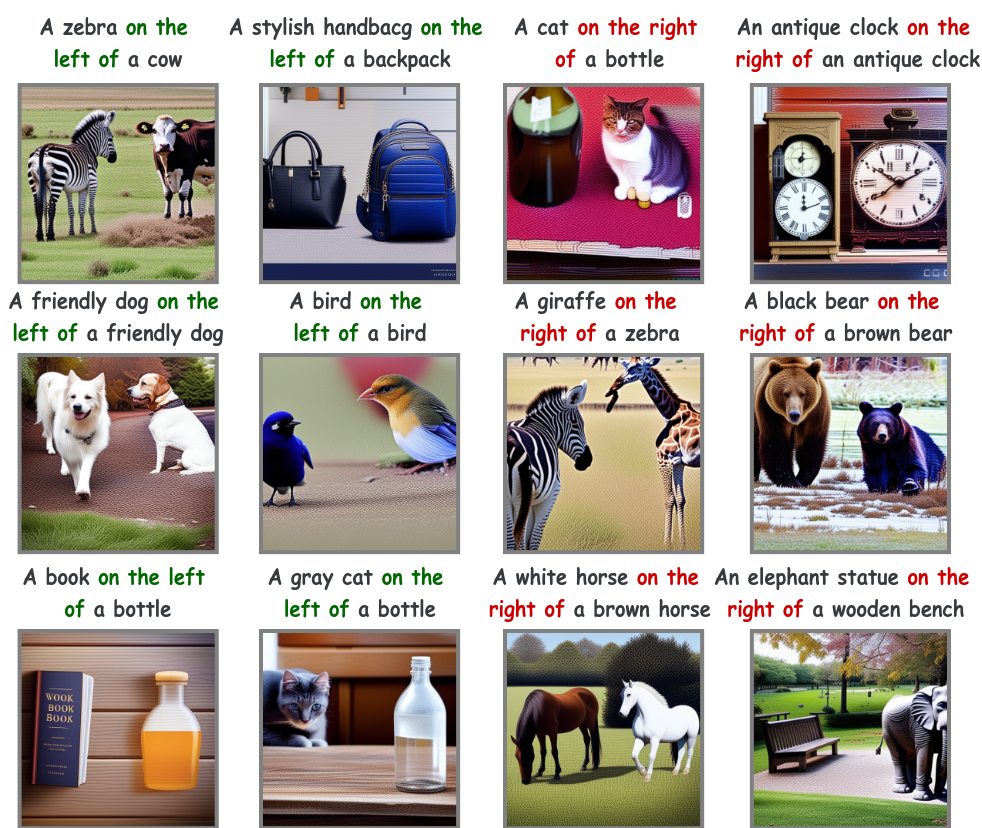}
    \caption{\textbf{Qualitative Evaluation.} Our model follows the spatial comprehension and attribute assignment in the the prompt.}
    \label{fig:qualitative_reuslts}
    \vspace{-5mm}
\end{figure}
Next, we qualitatively compare our \ours{} model against leading baselines as illustrated in Fig. \ref{fig:teaser}. As can be seen, for the text instruction ``A classic car on the right of a clock,'' all baseline models struggle to simultaneously generate both objects in the scene while our model adeptly generates the scene as described. For the text instruction ``A white bowl on the left of a brown bowl,'' we notice that while baseline models falter in precisely linking attributes to their respective objects, our \ours{} excels in accurately assigning these attributes and spatial comprehension. In regard with the spatial relationships between objects, both DALL-E 2 and CogView2 fall short in creating the spatial arrangement specified in the text instruction. However, our model generates images that faithfully reflect these descriptions. See Fig. \ref{fig:qualitative_reuslts} for more qualitative examples.

\subsection{Discussion \& Limitations}
\begin{figure}[t]
    \centering
    \includegraphics[width=\linewidth]{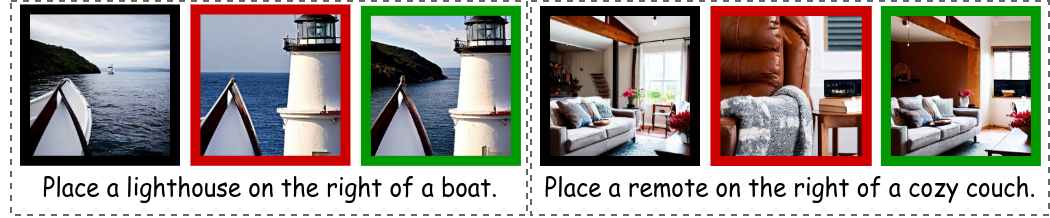}
    \caption{\textbf{Effectiveness of Background Retention:} Comparison of image synthesis with (green) and without (red) background retention energy function. As shown, our method effectively preserves background details of input image (black).}
    \label{fig:retention}
    \vspace{-5mm}
\end{figure}
\paragraph{Effectiveness of Background Retention.} This section evaluates the background retention energy function, as depicted in Fig. \ref{fig:retention}. In the figure, the image with a black outline represents the first image produced by our pipeline. The image with a red outline indicates the second image synthesized without the background retention energy function, while the one with a green outline demonstrates the second image synthesized with this function. Our proposed background retention energy function effectively transfers background information from the first to the second image, mitigating any negative alterations.

\paragraph{\ours{} Limitations.} Despite demonstrating improved spatial understanding, \ours{} is currently limited to generating only up to two objects with left and right relations in a scene, and it struggles with handling occlusions. These issues mainly arise from our dataset synthesis approach, focusing on `left 'and `right' object positioning. In our future work, we plan to expand the synthesis of more spatial relations. Furthermore, the use of GPT3-turbo, trained solely on text, lacks a deep grasp of geometrical aspects in scenes. Hence, a multimodal model, e.g. GPT4, might markedly improve image layout generation. These limitations highlight avenues for future research for which we aim to develop solutions that significantly boost our model's capabilities.

\section{Conclusion}
\label{sec:conclusion}
We have presented \ours, a pipeline that effectively enhances the capabilities of text-to-image models in comprehending spatial relationships and attribute assignments in text prompts. Our pipeline employs an iterative image editing process, which inserts objects into existing scenes based on their relative positions. This is made possible by our creation of a synthetic multimodal dataset for spatial reasoning, utilizing a frozen large language model and  a frozen layout-based diffusion model. Our experimental results validate the proficiency of \ours{} in interpreting the relationships between objects as described in text instructions and its effectiveness in attribute assignment. Moreover, despite being smaller size, our model demonstrates better performance compared to strong baselines.
{
    \small
    \bibliographystyle{ieeenat_fullname}
    \bibliography{main}

\begin{thebibliography}{39}
\providecommand{\natexlab}[1]{#1}
\providecommand{\url}[1]{\texttt{#1}}
\expandafter\ifx\csname urlstyle\endcsname\relax
  \providecommand{\doi}[1]{doi: #1}\else
  \providecommand{\doi}{doi: \begingroup \urlstyle{rm}\Url}\fi

\bibitem[Abadi et~al.(2016)Abadi, Agarwal, Barham, Brevdo, Chen, Citro, Corrado, Davis, Dean, Devin, et~al.]{abadi2016tensorflow}
Mart{\'\i}n Abadi, Ashish Agarwal, Paul Barham, Eugene Brevdo, Zhifeng Chen, Craig Citro, Greg~S Corrado, Andy Davis, Jeffrey Dean, Matthieu Devin, et~al.
\newblock Tensorflow: Large-scale machine learning on heterogeneous distributed systems.
\newblock \emph{arXiv preprint arXiv:1603.04467}, 2016.

\bibitem[Avrahami et~al.(2022)Avrahami, Lischinski, and Fried]{Avrahami_2022_CVPR}
Omri Avrahami, Dani Lischinski, and Ohad Fried.
\newblock Blended diffusion for text-driven editing of natural images.
\newblock In \emph{IEEE Conference on Computer Vision and Pattern Recognition}, 2022.

\bibitem[Brooks et~al.(2023)Brooks, Holynski, and Efros]{brooks2023instructpix2pix}
Tim Brooks, Aleksander Holynski, and Alexei~A. Efros.
\newblock Instructpix2pix: Learning to follow image editing instructions.
\newblock In \emph{IEEE Conference on Computer Vision and Pattern Recognition}, 2023.

\bibitem[Brown et~al.(2020)Brown, Mann, Ryder, Subbiah, Kaplan, Dhariwal, Neelakantan, Shyam, Sastry, Askell, et~al.]{brown2020language}
Tom Brown, Benjamin Mann, Nick Ryder, Melanie Subbiah, Jared~D Kaplan, Prafulla Dhariwal, Arvind Neelakantan, Pranav Shyam, Girish Sastry, Amanda Askell, et~al.
\newblock Language models are few-shot learners.
\newblock \emph{Advances on Neural Information Processing Systems}, 2020.

\bibitem[Couairon et~al.(2023)Couairon, Verbeek, Schwenk, and Cord]{couairon2023diffedit}
Guillaume Couairon, Jakob Verbeek, Holger Schwenk, and Matthieu Cord.
\newblock Diffedit: Diffusion-based semantic image editing with mask guidance.
\newblock In \emph{International Conference on Learning Representations}, 2023.

\bibitem[Dayma et~al.(2021)Dayma, Patil, Cuenca, Saifullah, Abraham, Lê~Khac, Melas, and Ghosh]{Dayma_DALL·E_Mini_2021}
Boris Dayma, Suraj Patil, Pedro Cuenca, Khalid Saifullah, Tanishq Abraham, Phúc Lê~Khac, Luke Melas, and Ritobrata Ghosh.
\newblock Dall·e mini, 2021.

\bibitem[Dhariwal and Nichol(2021)]{dhariwal2021diffusion}
Prafulla Dhariwal and Alexander~Quinn Nichol.
\newblock Diffusion models beat {GAN}s on image synthesis.
\newblock In \emph{Advances on Neural Information Processing Systems}, 2021.

\bibitem[Ding et~al.(2022)Ding, Zheng, Hong, and Tang]{ding2022cogview2}
Ming Ding, Wendi Zheng, Wenyi Hong, and Jie Tang.
\newblock Cogview2: Faster and better text-to-image generation via hierarchical transformers.
\newblock \emph{Advances on Neural Information Processing Systems}, 2022.

\bibitem[Gokhale et~al.(2022)Gokhale, Palangi, Nushi, Vineet, Horvitz, Kamar, Baral, and Yang]{gokhale2022benchmarking}
Tejas Gokhale, Hamid Palangi, Besmira Nushi, Vibhav Vineet, Eric Horvitz, Ece Kamar, Chitta Baral, and Yezhou Yang.
\newblock Benchmarking spatial relationships in text-to-image generation.
\newblock \emph{arXiv preprint arXiv:2212.10015}, 2022.

\bibitem[Hertz et~al.(2022)Hertz, Mokady, Tenenbaum, Aberman, Pritch, and Cohen-or]{hertz2022prompt}
Amir Hertz, Ron Mokady, Jay Tenenbaum, Kfir Aberman, Yael Pritch, and Daniel Cohen-or.
\newblock Prompt-to-prompt image editing with cross-attention control.
\newblock In \emph{International Conference on Learning Representations}, 2022.

\bibitem[Ho and Salimans(2022)]{ho2022classifier}
Jonathan Ho and Tim Salimans.
\newblock Classifier-free diffusion guidance.
\newblock \emph{arXiv preprint arXiv:2207.12598}, 2022.

\bibitem[Ho et~al.(2020)Ho, Jain, and Abbeel]{Ho2020Denoising}
Jonathan Ho, Ajay Jain, and Pieter Abbeel.
\newblock Denoising diffusion probabilistic models.
\newblock In \emph{Advances on Neural Information Processing Systems}, 2020.

\bibitem[Ho et~al.(2022)Ho, Saharia, Chan, Fleet, Norouzi, and Salimans]{ho2022cascaded}
Jonathan Ho, Chitwan Saharia, William Chan, David~J Fleet, Mohammad Norouzi, and Tim Salimans.
\newblock Cascaded diffusion models for high fidelity image generation.
\newblock \emph{The Journal of Machine Learning Research}, 2022.

\bibitem[Hu et~al.(2022)Hu, Shen, Wallis, Allen-Zhu, Li, Wang, Wang, and Chen]{hu2021lora}
Edward~J Hu, Yelong Shen, Phillip Wallis, Zeyuan Allen-Zhu, Yuanzhi Li, Shean Wang, Lu Wang, and Weizhu Chen.
\newblock Lora: Low-rank adaptation of large language models.
\newblock In \emph{International Conference on Learning Representations}, 2022.

\bibitem[Huberman-Spiegelglas et~al.(2023)Huberman-Spiegelglas, Kulikov, and Michaeli]{HubermanSpiegelglas2023}
Inbar Huberman-Spiegelglas, Vladimir Kulikov, and Tomer Michaeli.
\newblock An edit friendly ddpm noise space: Inversion and manipulations.
\newblock \emph{arXiv preprint arXiv:2304.06140}, 2023.

\bibitem[Kingma and Ba(2016)]{kingma2014adam}
Diederik~P Kingma and Jimmy Ba.
\newblock Adam: A method for stochastic optimization.
\newblock \emph{International Conference on Learning Representations}, 2016.

\bibitem[Li et~al.(2023)Li, Liu, Wu, Mu, Yang, Gao, Li, and Lee]{li2023gligen}
Yuheng Li, Haotian Liu, Qingyang Wu, Fangzhou Mu, Jianwei Yang, Jianfeng Gao, Chunyuan Li, and Yong~Jae Lee.
\newblock Gligen: Open-set grounded text-to-image generation.
\newblock In \emph{IEEE Conference on Computer Vision and Pattern Recognition}, 2023.

\bibitem[Lian et~al.(2023)Lian, Li, Yala, and Darrell]{lian2023llmgrounded}
Long Lian, Boyi Li, Adam Yala, and Trevor Darrell.
\newblock Llm-grounded diffusion: Enhancing prompt understanding of text-to-image diffusion models with large language models.
\newblock \emph{arXiv preprint arXiv:2305.13655}, 2023.

\bibitem[Lin et~al.(2014)Lin, Maire, Belongie, Hays, Perona, Ramanan, Doll{\'a}r, and Zitnick]{lin2014microsoft}
Tsung-Yi Lin, Michael Maire, Serge Belongie, James Hays, Pietro Perona, Deva Ramanan, Piotr Doll{\'a}r, and C~Lawrence Zitnick.
\newblock Microsoft coco: Common objects in context.
\newblock In \emph{European Conference on Computer Vision}, 2014.

\bibitem[Liu et~al.(2022)Liu, Li, Du, Torralba, and Tenenbaum]{liu2022compositional}
Nan Liu, Shuang Li, Yilun Du, Antonio Torralba, and Joshua~B Tenenbaum.
\newblock Compositional visual generation with composable diffusion models.
\newblock In \emph{European Conference on Computer Vision}, 2022.

\bibitem[Minderer et~al.(2022)Minderer, Gritsenko, Stone, Neumann, Weissenborn, Dosovitskiy, Mahendran, Arnab, Dehghani, Shen, et~al.]{minderer2022simple}
Matthias Minderer, Alexey Gritsenko, Austin Stone, Maxim Neumann, Dirk Weissenborn, Alexey Dosovitskiy, Aravindh Mahendran, Anurag Arnab, Mostafa Dehghani, Zhuoran Shen, et~al.
\newblock Simple open-vocabulary object detection.
\newblock In \emph{European Conference on Computer Vision}, 2022.

\bibitem[Mokady et~al.(2023)Mokady, Hertz, Aberman, Pritch, and Cohen-Or]{mokady2023null}
Ron Mokady, Amir Hertz, Kfir Aberman, Yael Pritch, and Daniel Cohen-Or.
\newblock Null-text inversion for editing real images using guided diffusion models.
\newblock In \emph{IEEE Conference on Computer Vision and Pattern Recognition}, 2023.

\bibitem[Nichol et~al.(2022)Nichol, Dhariwal, Ramesh, Shyam, Mishkin, Mcgrew, Sutskever, and Chen]{nichol2022glide}
Alexander~Quinn Nichol, Prafulla Dhariwal, Aditya Ramesh, Pranav Shyam, Pamela Mishkin, Bob Mcgrew, Ilya Sutskever, and Mark Chen.
\newblock {GLIDE}: Towards photorealistic image generation and editing with text-guided diffusion models.
\newblock In \emph{International Conference on Learning Representations}, 2022.

\bibitem[Peebles and Xie(2023)]{peebles2023scalable}
William Peebles and Saining Xie.
\newblock Scalable diffusion models with transformers.
\newblock In \emph{IEEE Conference on Computer Vision and Pattern Recognition}, 2023.

\bibitem[Podell et~al.(2023)Podell, English, Lacey, Blattmann, Dockhorn, M{\"u}ller, Penna, and Rombach]{podell2023sdxl}
Dustin Podell, Zion English, Kyle Lacey, Andreas Blattmann, Tim Dockhorn, Jonas M{\"u}ller, Joe Penna, and Robin Rombach.
\newblock Sdxl: Improving latent diffusion models for high-resolution image synthesis.
\newblock \emph{arXiv preprint arXiv:2307.01952}, 2023.

\bibitem[Radford et~al.(2021)Radford, Kim, Hallacy, Ramesh, Goh, Agarwal, Sastry, Askell, Mishkin, Clark, et~al.]{radford2021learning}
Alec Radford, Jong~Wook Kim, Chris Hallacy, Aditya Ramesh, Gabriel Goh, Sandhini Agarwal, Girish Sastry, Amanda Askell, Pamela Mishkin, Jack Clark, et~al.
\newblock Learning transferable visual models from natural language supervision.
\newblock In \emph{International Conference on Machine Learning}, 2021.

\bibitem[Ramesh et~al.(2022)Ramesh, Dhariwal, Nichol, Chu, and Chen]{ramesh2022hierarchical}
Aditya Ramesh, Prafulla Dhariwal, Alex Nichol, Casey Chu, and Mark Chen.
\newblock Hierarchical text-conditional image generation with clip latents.
\newblock \emph{arXiv preprint arXiv:2204.06125}, 2022.

\bibitem[Rombach et~al.(2022)Rombach, Blattmann, Lorenz, Esser, and Ommer]{Rombach_2022_CVPR}
Robin Rombach, Andreas Blattmann, Dominik Lorenz, Patrick Esser, and Bj\"orn Ommer.
\newblock High-resolution image synthesis with latent diffusion models.
\newblock In \emph{IEEE Conference on Computer Vision and Pattern Recognition}, 2022.

\bibitem[Ronneberger et~al.(2015)Ronneberger, Fischer, and Brox]{ronneberger2015u}
Olaf Ronneberger, Philipp Fischer, and Thomas Brox.
\newblock U-net: Convolutional networks for biomedical image segmentation.
\newblock In \emph{International Conference on Medical Image Computing and Computer-Assisted Intervention}, 2015.

\bibitem[Ruiz et~al.(2023)Ruiz, Li, Jampani, Pritch, Rubinstein, and Aberman]{ruiz2023dreambooth}
Nataniel Ruiz, Yuanzhen Li, Varun Jampani, Yael Pritch, Michael Rubinstein, and Kfir Aberman.
\newblock Dreambooth: Fine tuning text-to-image diffusion models for subject-driven generation.
\newblock In \emph{IEEE Conference on Computer Vision and Pattern Recognition}, 2023.

\bibitem[Saharia et~al.(2022)Saharia, Chan, Saxena, Li, Whang, Denton, Ghasemipour, Gontijo~Lopes, Karagol~Ayan, Salimans, et~al.]{saharia2022photorealistic}
Chitwan Saharia, William Chan, Saurabh Saxena, Lala Li, Jay Whang, Emily~L Denton, Kamyar Ghasemipour, Raphael Gontijo~Lopes, Burcu Karagol~Ayan, Tim Salimans, et~al.
\newblock Photorealistic text-to-image diffusion models with deep language understanding.
\newblock \emph{Advances on Neural Information Processing Systems}, 2022.

\bibitem[Saharia et~al.(2023)Saharia, Ho, Chan, Salimans, Fleet, and Norouzi]{saharia2023Image}
Chitwan Saharia, Jonathan Ho, William Chan, Tim Salimans, David~J. Fleet, and Mohammad Norouzi.
\newblock Image super-resolution via iterative refinement.
\newblock \emph{PAMI}, 2023.

\bibitem[Sohl-Dickstein et~al.(2015)Sohl-Dickstein, Weiss, Maheswaranathan, and Ganguli]{Sohl2015Deep}
Jascha Sohl-Dickstein, Eric Weiss, Niru Maheswaranathan, and Surya Ganguli.
\newblock Deep unsupervised learning using nonequilibrium thermodynamics.
\newblock In \emph{International Conference on Machine Learning}, 2015.

\bibitem[Song et~al.(2020)Song, Meng, and Ermon]{song2020denoising}
Jiaming Song, Chenlin Meng, and Stefano Ermon.
\newblock Denoising diffusion implicit models.
\newblock In \emph{International Conference on Learning Representations}, 2020.

\bibitem[Song et~al.(2021)Song, Meng, and Ermon]{song2021denoising}
Jiaming Song, Chenlin Meng, and Stefano Ermon.
\newblock Denoising diffusion implicit models.
\newblock In \emph{International Conference on Learning Representations}, 2021.

\bibitem[Song and Ermon(2019)]{Song2019Generative}
Yang Song and Stefano Ermon.
\newblock Generative modeling by estimating gradients of the data distribution.
\newblock In \emph{Advances on Neural Information Processing Systems}, 2019.

\bibitem[Vaswani et~al.(2017)Vaswani, Shazeer, Parmar, Uszkoreit, Jones, Gomez, Kaiser, and Polosukhin]{vaswani2017attention}
Ashish Vaswani, Noam Shazeer, Niki Parmar, Jakob Uszkoreit, Llion Jones, Aidan~N Gomez, \L~ukasz Kaiser, and Illia Polosukhin.
\newblock Attention is all you need.
\newblock In \emph{Advances on Neural Information Processing Systems}, 2017.

\bibitem[von Platen et~al.(2022)von Platen, Patil, Lozhkov, Cuenca, Lambert, Rasul, Davaadorj, and Wolf]{von2022diffusers}
Patrick von Platen, Suraj Patil, Anton Lozhkov, Pedro Cuenca, Nathan Lambert, Kashif Rasul, Mishig Davaadorj, and Thomas Wolf.
\newblock Diffusers: State-of-the-art diffusion models, 2022.

\bibitem[Zhang et~al.(2023)Zhang, Zhang, Vineet, Joshi, and Wang]{zhang2023controllable}
Tianjun Zhang, Yi Zhang, Vibhav Vineet, Neel Joshi, and Xin Wang.
\newblock Controllable text-to-image generation with gpt-4.
\newblock \emph{arXiv preprint arXiv:2305.18583}, 2023.

\end{thebibliography}
}

\clearpage
\setcounter{page}{1}
\maketitlesupplementary

\section{Layout Generation}
In this section, we utilize LLMs to produce an image layout based on the caption generated. Each constructed layout consists of two primary components: instance-level annotations using a collection of captioned bounding boxes and a background prompt. Instance-level annotations are defined as \texttt{(caption,  [x, y, w, h])}, where \texttt{[x, y, w, h]} specifies the bounding box location of each object and \texttt{caption} determines the content of each bounding box. However, we found that decomposing foreground objects from background is challenging due to the hallucination of LLMs. To address this, similar to previous step, we stick to the in-context learning capabilities of LLMs and precisely control the layout generation through two key designs. First, we ensure that every object instance is associated to a single bounding box. Second, any foreground objects detailed within the boxes should not exist in the background prompt. This ensures that our layout-grounded image generator exclusively manages all foreground objects. Check right column for more detail on the prompt format.

\section{Hyperparameters Details}
\label{sec:more_impl_details}
\begin{table}[h]
  \centering
    \resizebox{0.9\linewidth}{!}
    {\setlength\tabcolsep{4pt}
    \begin{tabular}{ll}
    \toprule
    \textbf{Hyperparameters} & \textbf{Fine-tuning} \\
    \midrule
    LM choice & GPT3.5-turbo \\
    Layout-based Diffusion Model  & GLIGEN \\
    Optimizer  & AdamW  \\
    Learning rate   & $1e-4$\\
    LR scheduler  & Linear  \\
    Batch size  & $8$  \\
    Iterations  & $12{,}000$  \\
    Warm-up steps  & $0$  \\
    Images size & $512 \times 512$ \\
    GPUs & $4$ A$5000$ \\
    \bottomrule
  \end{tabular}
  }
  \caption{
  List of hyperparameters used to reproduce the experimental results in the paper for the fine-tuning stage of our \ours{}.
  }
  \label{tab:hyper}
\end{table}
We provide the detailed hyperparameters used for the fine-tuning stage of our \ours{} in Table \ref{tab:hyper}. Our \ours{} is trained using mixed-precision with Bfloat16. In the adaptation stage, we use a batch size of $8$ over $12{,}000$ iterations and 4 A5000 GPUs. Furthermore, we use the AdamW optimizer with a learning rate of $1e$-$4$ and a linear learning rate schedular. Our \ours{} is initialized using the pre-trained weights of InstructPix2Pix, chosen for its superior performance in global edits, such as texture and lighting adjustments. We fine-tune our model using LoRA while keeping the pre-trained weights fixed.

% \vfill\null
% \columnbreak
\begin{tcolorbox}[colback=red!5!white,colframe=red!75!black, width=0.5\textwidth, size=small,title=Layout Generation Prompt with ICL]
\noindent \textbf{[Task Description]:} You are a smart bounding box generator. I will provide you with a caption for a photo, image, or painting. Your task is to generate the bounding boxes for the objects mentioned in the caption, along with a background prompt describing the scene. If needed, you can make reasonable guesses. The images are of size 512x512, and the bounding boxes should not overlap or go beyond the image boundaries. Each bounding box should be in the format of (object name,  [x, y, width, height]), with the constraint that width and height are both less than 350. Please refer to the example below for the desired format.
\newline
\textbf{[Caption]:} A watercolor painting of two pandas eating bamboo in a forest.
\newline
\textbf{[Objects]:} [('a panda eating bambooo', [30, 171, 212, 226]), ('a panda eating bambooo', [264, 173, 222, 221])]
\newline
\textbf{[Background prompt]:} A watercolor painting of a forest
\newline
\newline
\textbf{[Caption]:} An oil painting of a pink dolphin jumping on the left of a steamboat on the sea.
\newline
\textbf{[Objects]:} [('a steamboat', [232, 225, 257, 149]), ('a jumping pink dolphin', [21, 249, 189, 123])]
\newline
\textbf{[Background prompt]:} An oil painting of the sea
\newline
\newline
\textbf{[Caption]:} A realistic image of a cat playing with a dog in a park with flowers.
\newline
\textbf{[Objects]:} [('a playful cat', [51, 67, 271, 324]), ('a playful dog', [302, 119, 211, 228])]
\newline
\textbf{[Background prompt]:} A realistic image of a park with flowers
\newline
\newline
\textbf{[Caption]:} A realistic photograph of a scene with a dog on the left and a tree on the right.
\newline
\textbf{[Objects]:} [('a dog', [3, 122, 212, 250]), ('a tree', [287, 31, 220, 341])]
\newline
\textbf{[Background prompt]:}: A realistic photograph of a scene
\newline
\newline
\textbf{[Caption]:} A realistic photo of a mountain with a lake on the left and a tree on the right.
\newline
\textbf{[Objects]:} $<\text{LLM-generated}>$
\end{tcolorbox}

We provide the detailed hyperparameters for the generation of captions and layout used in the image synthesis section of our \ours{} in Table \ref{tab:hyper-llm}. 
% request_params = {
%             "temperature": 1,
%             "max_tokens": args.max_output_len,
%             "top_p": 0.5,
%             "frequency_penalty": 0,
%             "presence_penalty": 0,
%             "n": 1,
%             "stop": stop_character,
%         }

\begin{table}[h]
  \centering
    \resizebox{\linewidth}{!}
    {\setlength\tabcolsep{4pt}
    \begin{tabular}{lll}
    \toprule
    \textbf{Hyperparameters} & \textbf{Caption Generation} & \textbf{Layout Generation} \\
    \midrule
    Temperature       & $1$   & $1$ \\
    Max Tokens        & $100$ & $100$ \\
    $\text{Top}_{p}$             & $0.5$ & $0.5$ \\
    Frequency Penalty & $0$   & $0$ \\
    Presence Penalty  & $0$   & $0$ \\
    Number of samples & $100$ & $100$ \\
    Stop Character    & `.'   & `$\backslash{n}\backslash{n}$' \\
    \bottomrule
  \end{tabular}
  }
  \caption{
  List of hyperparameters utilized for creating captions and layouts with GPT-3.5-turbo in the dataset synthesis process of our model, referred to as \ours{}.
  }
  \label{tab:hyper-llm}
\end{table}

\section{Image Enhancement}
\label{sec:image_enhancement}
\begin{figure}[h]
    \centering
    \includegraphics[width=\linewidth]{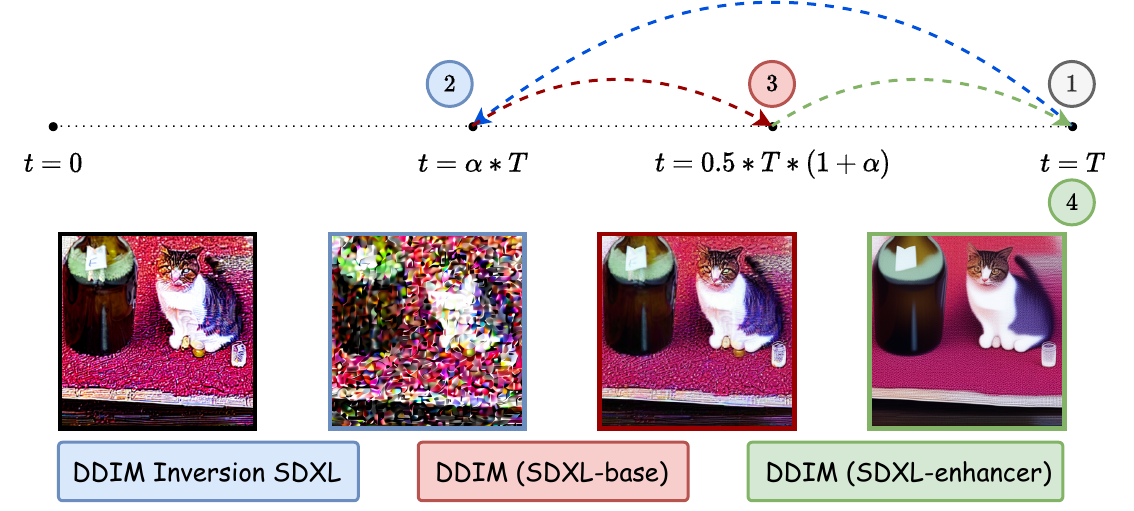}
    \caption{Overview of our image enhancement pipeline integrating pre-trained Stable Diffusion XL (base and enhancer model) with DDIM inversion to improve the \ours{} outputs.}
    \label{fig:enhancement}
\end{figure}
In this section we propose an efficient pipeline, utilizing Stable Diffusion XL model in tandem with DDIM \cite{song2020denoising} to enhance outputs from \ours{} as depicted in Fig. \ref{fig:enhancement}. The process involves three key phases: 
\begin{enumerate}
    \item \textbf{Noise addition using DDIM inversion (blue phase):} We add noise to the generated image using DDIM inversion, progressing from step $T$ to $t = \alpha \times T$.
    \item \textbf{Noise reduction using DDIM (red phase):} From step $t = \alpha \times T$ to $t = 0.5 \times T \times (1 + \alpha)$, we systematically remove the noise and recover the generated image by leveraging DDIM process in conjunction with the SDXL-base model.
    \item \textbf{Image enhancement (green phase):} We enhance image quality by employing the SDXL-enhancer model. Improvement occurs through the DDIM process, from step $t = 0.5 \times T \times (1 + \alpha)$ back to $T$.
\end{enumerate}
Each phase is designed to incrementally refine the image, balancing the correctness of the relationship between objects and efficiency in the enhancement process by leveraging pretrained SDXL models. In this process, we employ a total of 100 diffusion steps, denoted as 
$T$, and set the parameter 
$\alpha=0.7$. Regarding the pretrained models employed in this section, we utilize the SDXL-base and SDXL-enhancer models, which are provided in the Diffuser library. 

\section{Dataset Synthesis}
\begin{figure*}[t]
    \centering
    \includegraphics[width=\textwidth]{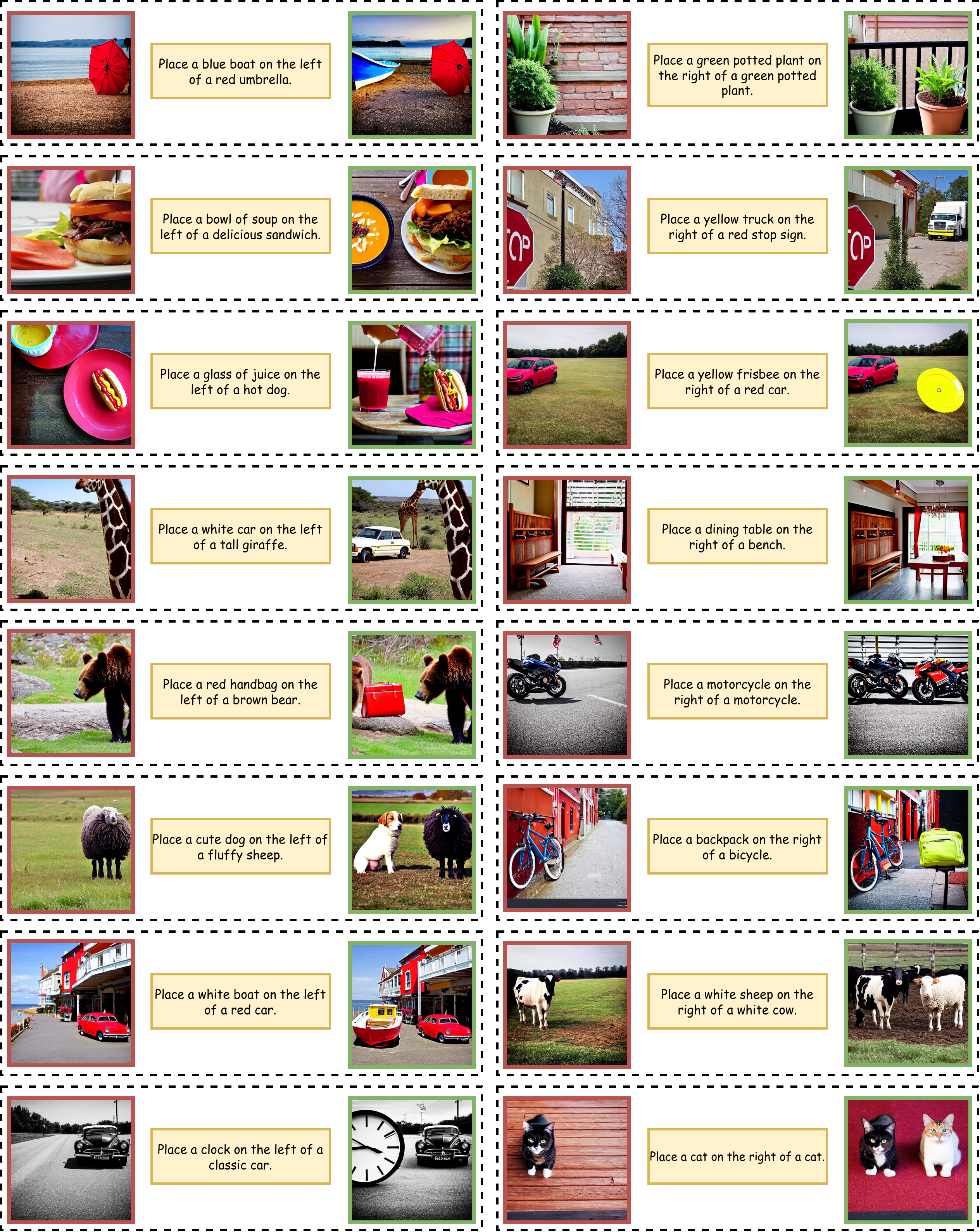}
    \caption{\textbf{Data synthesis.} This figure illustrates the training triplets utilized to train our model, \ours{}. Each triplet consists of three components: image conditioning (left), text instruction (middle), and the output image (right), which is predicted by \ours{}.}
    \label{fig:synthesized_dataset}
    \vspace{-5mm}
\end{figure*}

In this section, we provide several synthesized dataset that are used to finetune our \ours{} for spatial comprehension and attribute assignment tasks in Fig. \ref{fig:synthesized_dataset}. This figure illustrates the training triplets utilized to train our \ours{}. Each triplet consists of three components: image conditioning (left), text instruction (middle), and the output image (right), which is predicted by \ours{}. These samples are synthesized by leveraging a frozen LLM, e.g. GPT3-turbo, and a frozen layout-based diffusion model, e.g. LDM.

\begin{figure*}[t]
    \centering
    \includegraphics[width=\textwidth]{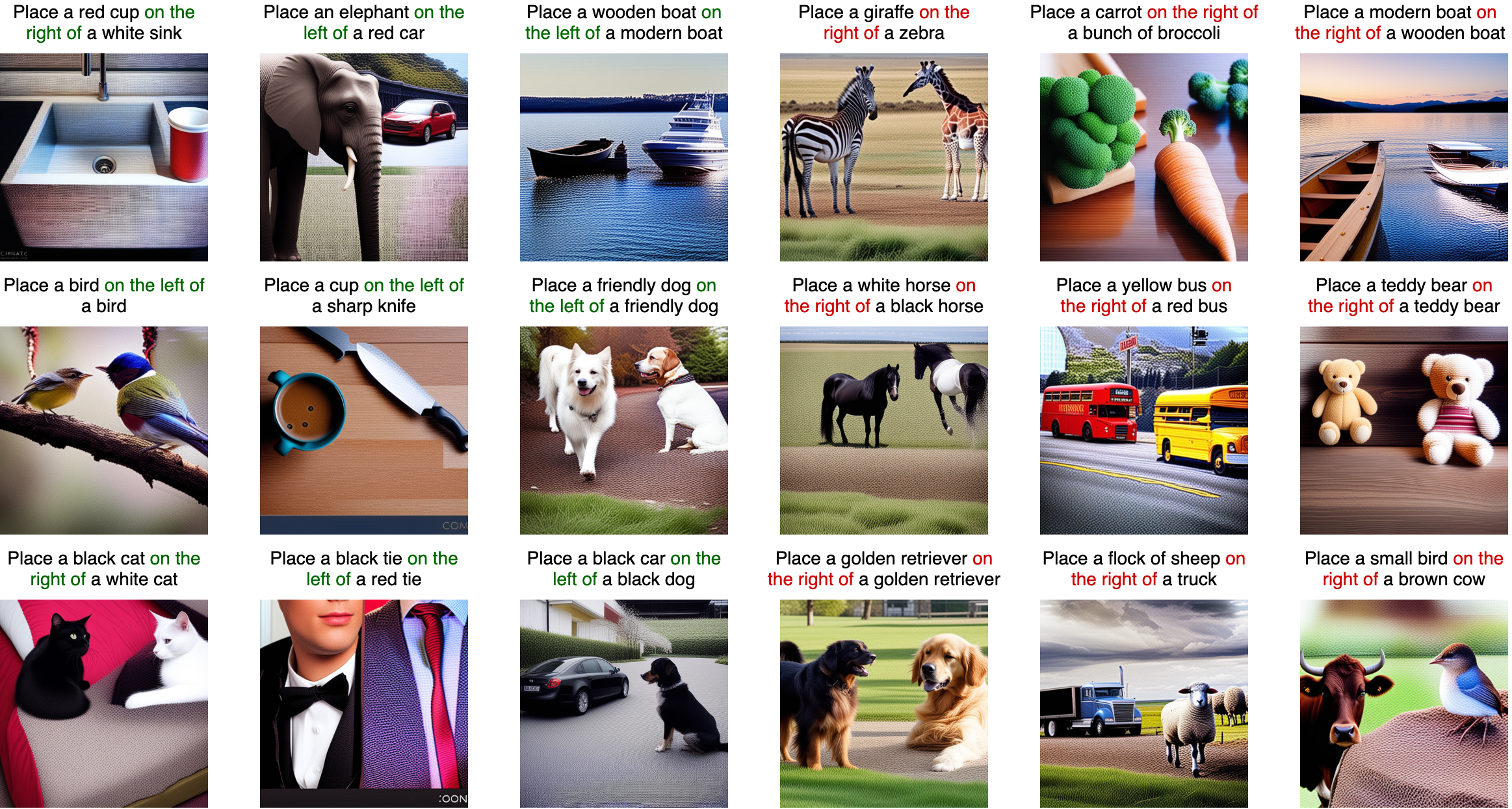}
    \caption{\textbf{Qualitative Evaluation.} Our model follows the spatial comprehension and attribute assignment in the the prompt.}
    \label{fig:more_eval}
    \vspace{-5mm}
\end{figure*}

\end{document}